%% file: IEEE.tex
\documentclass[lettersize,journal]{IEEEtran}

\input{assets/documents/00packages.tex}
\usepackage[numbers]{natbib}
\let\oldcitep\citep
\renewcommand{\citep}[1]{\mbox{\oldcitep{#1}}}
\let\oldcitet\citet
\renewcommand{\citet}[1]{\mbox{\oldcitet{#1}}}
\let\oldcite\cite
\renewcommand{\cite}[1]{\mbox{\oldcite{#1}}}

\begin{document}

\title{Dynamic Collaborative Material Distribution System for Intelligent Robots In Smart Manufacturing}

\author{Ziren Xiao,
        Ruxin Xiao,
        Chang Liu,
        Xinheng Wang

\thanks{This research received partial support from the XJTLU Research Development Funding under projects RDF-19-01-14 and RDF-20-01-15, and the National Natural Science Foundation of China (NSFC) under grant 52175030. (\emph{Corresponding author: Xinheng Wang})}
\thanks{Ziren Xiao, Ruxin Xiao, Chang Liu, Sixuan Duan and Xinheng Wang are with the School of Advanced Technology, Xi'an Jiaotong-Liverpool University, Suzhou, 215123, China, E-mail: ziren.xiao20@alumni.xjtlu.edu.cn;
ruxin.xiao21@student.xjtlu.edu.cn;chang.liu17@student.xjtlu.edu.cn;
xinheng.wang@xjtlu.edu.cn. }

}

\markboth{IEEE Robotics and Automation Letters}{Dynamic Collaborative Material Distribution System for Intelligent Robots In Smart Manufacturing}

\maketitle

\input{assets/documents/01abstract.tex}


\IEEEpeerreviewmaketitle

\input{assets/documents/02introduction_old}

\input{assets/documents/03relatedwork}

\input{assets/documents/04methodology}

\input{assets/documents/05evaluation}

\input{assets/documents/06conclusion}


\bibliographystyle{IEEEtranN}
\bibliography{references}

\end{document}

%% file: assets/documents/00packages.tex
\usepackage{lineno,hyperref}
\usepackage{amsmath,amssymb,amsfonts}
\usepackage{graphicx}
\usepackage{textcomp}
\usepackage{xcolor}
\usepackage{algorithm}
\usepackage{algorithmic}
\usepackage{wrapfig}
\usepackage{booktabs}
\usepackage{multirow}
\usepackage{multicol}
\usepackage{subcaption}
\usepackage[final]{changes} 

%% file: assets/documents/01abstract.tex
\begin{abstract}
The collaboration and interaction of multiple robots have become integral aspects of smart manufacturing. Effective planning and management play a crucial role in achieving energy savings and minimising overall costs. This paper addresses the real-time Dynamic Multiple Sources to Single Destination (DMS-SD) navigation problem, particularly with a material distribution case for multiple intelligent robots in smart manufacturing. Enumerated solutions, such as in \cite{xiao2022efficient}, tackle the problem by generating as many optimal or near-optimal solutions as possible but do not learn patterns from the previous experience, whereas the method in \cite{xiao2023collaborative} only uses limited information from the earlier trajectories. Consequently, these methods may take a considerable amount of time to compute results on large maps, rendering real-time operations impractical. To overcome this challenge, we propose a lightweight Deep Reinforcement Learning (DRL) method to address the DMS-SD problem. The proposed DRL method can be efficiently trained and rapidly converges to the optimal solution using the designed target-guided reward function. A well-trained DRL model significantly reduces the computation time for the next movement to a millisecond level, which improves the time up to 100 times in our experiments compared to the enumerated solutions. Moreover, the trained DRL model can be easily deployed on lightweight devices in smart manufacturing, such as Internet of Things devices and mobile phones, which only require limited computational resources.
\end{abstract}

%% file: assets/documents/02introduction_old.tex
\section{Introduction}

The importance of swarm \replaced{robots}{robotics} in modern industry is dramatically increasing. In past years, pervasive smart manufacturing is arguably the most revolutionary trend in modern industry. In the prospect of smart manufacturing, modern industry is evolving into a more digitised, autonomous, and adaptive form. In this background, swarm \replaced{robots}{robotics}, including the community of homogeneous or heterogeneous collaborative smart robots, tend to be more critical than ever before. Conventionally, the deployment of swarm \replaced{robots}{robotics} relies on internal algorithms and an interconnected network to complete regular tasks in the industry, such as material delivery, safety monitoring, and cleaning. Following the development of artificial intelligence and the industrial internet of things, swarm \replaced{robots}{robotics} tend to work in more complicated environments and handle more challenging tasks. 

\begin{figure}
    \centering
    \includegraphics[width=0.5\textwidth]{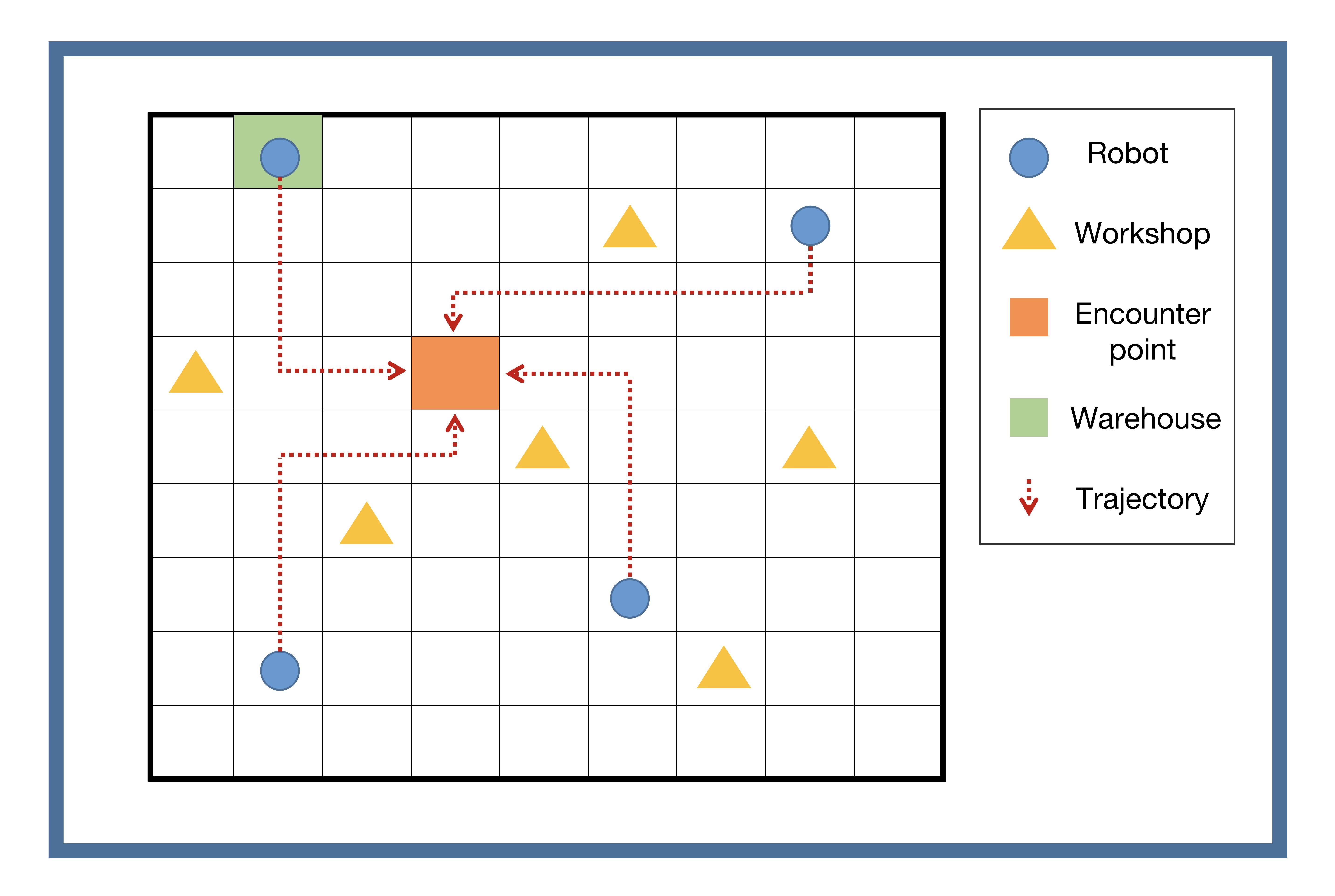}
\caption{A grid-based sketch of DMS-SD in factory. There are three multi-function robots in different locations around workshops and one loading robot in a warehouse. They need to meet at one point and distribute materials to workshops. Everyone travels a similar and shortest distance and time within the factory layout to ensure effective production.}
\label{fig:example}
\end{figure}

The introduction of swarm robots to factories with automated manufacturing has recently led to new challenges. Among these challenges, we particularly focus on path planning for dynamic multi-source to single destination (DMS-SD) navigation, where the destination can dynamically be changed while robots are moving~\cite{xiao2023collaborative}. In modern industry, DMS-SD is a ubiquitous problem \deleted{that appears} in many tasks of swarm \replaced{robots}{robotics}. We present a DMS-SD problem of material transportation in Fig.~\ref{fig:example}, where heterogeneous robots take on different duties in the factory under one control system. Within these swarm \replaced{robots}{robotics}, one robot with its specially designed structure is dedicated to material withdrawal from the warehouse. Then, the robot distributes materials to other robots, \added{delivering} collected materials to their designated workshops. Such task distribution aims to improve overall transportation efficiency by reducing the overall travelling distance and prohibiting congestion at the warehouse. In this condition, the control system of these swarm \replaced{robots}{robotics} must determine a destination as the rally point that the swarm robots meet with the minimised overall travelling distance. Moreover, due to the fluctuating industrial environment, the control system must adjust the destination and re-plan the path if there are accidents or congestion in the initial path. Therefore, this scenario forms a DMS-SD problem of material delivery with swarm \replaced{robots}{robotics}. 

A key challenge of fast response to requests still remains in the previous DMS-SD solver~\cite{xiao2023collaborative}. In order to address this challenge\added{,} we first design a \deleted{framework that integrates the previously modified Dijkstra algorithm (MDA)~\cite{xiao2022efficient} and a} Deep Reinforcement Learning (DRL)-based \added{framework using} Proximal Policy Optimisation (PPO)~\cite{schulman2017proximal} \added{with the masking mechanism~\cite{huang2020closer} to achieve better performance on collision avoidance to both static and dynamic obstacles}. \deleted{The MDA stores the path with the shortest distance from one point to another, while the DRL algorithm can learn patterns from previous experiences. After the training, a well-trained agent can be directly used to find the final destination without cooperating with the MDA.} Secondly, a new reward function is designed for the DRL agent. This reward function provides feedback on the input action to the learning agent, leading to a simple learning process for the DRL agent\added{,} which achieves higher performance and accuracy than other existing multi-agent solvers or simple distance-based reward functions. Through extensive evaluations, our proposed method can quickly improve the steps taken to the destination for each robot to a near-optimal value and then converge to an optimal solution after a few further iterations of fine tunings, which makes real-time operations of robots possible. Thirdly, a few measures are adopted to enhance the optimisation of the PPO algorithm, including adopting a homogeneous neural network model for all robots instead of training an independent model for each robot and collaboration of all robots to evaluate the global state to make a prediction rather than being `selfish’. By employing all these measures, the calculation time could be reduced to a millisecond level, omitting frequent message exchanges between robots. This ensures a fast response to changes and real-time operations, which addresses the challenge remaining for dynamic navigation.

The main contributions of this paper are listed as follows:
\begin{itemize}
    \item We propose a highly effective and practical model-free method to solve \added{the} DMS-SD problem. This method converges fast to the optimal and shows a higher level of performance compared with classical algorithms.
    \item We propose a novel reward function that penalties the wrong movement and rewards the movement toward the common destination computed by the \replaced{centroid of robots' positions}{modified Dijkstra's algorithm}, which guides \replaced{them}{users} moving toward the target.
    \item We apply a training strategy where all \replaced{robots}{users} share the same DRL model instead of holding an individual model for each \replaced{robot}{user}, so the agent can focus on training a single neural network. This is particularly useful when the patterns or behaviours of each \replaced{robot}{user} are very close.
\end{itemize}



%% file: assets/documents/03relatedwork.tex
\section{Related Work}
\begin{table*}[ht!]
\centering
\resizebox{\textwidth}{!}{%
\begin{tabular}{lllllllll}
Source & DRL Algorithm & Agent Architecture & Interaction & Obstacle & Object & Reward Objective & Goal Type \\ \hline
\cite{han2020cooperative} & PPO & Decentralised & Cooperation & Dynamic & Robot & Distance to Target & Static, Multiple \\
\cite{long2018towards} & PPO & Decentralised & Competition & Dynamic & Robot & Collision Avoidance & Static, Multiple \\
\cite{liu2021visuomotor} & PPO & Decentralised & Cooperation & Static & Robot & Correct Path & Static, Multiple \\
\cite{mohseni2019interaction} & Actor-Critic & Decentralised & Competition & Dynamic & Robot & Collision Avoidance & Static, Multiple \\
\cite{sivanathan2020decentralized} & PPO & Decentralised & Cooperation & Dynamic & Robot & Correct Path & Static, Multiple \\
\cite{xue2023multi} & Recurrent Actor-Critic & CTDE & Cooperation & Static & UAV & Correct Path & Static, Multiple \\
\cite{yang2020autonomous} & Double DQN & Decentralised & Cooperation & Dynamic & UAV & Correct Path & Static, Single \\
\cite{moon2021deep} & Dueling DQN & Decentralised & Cooperation & None & UAV & Correct Path & Dynamic, Multiple \\
\cite{huang2022vision} & SAC+AE & Decentralised & None & Dynamic & UAV & Correct Path & Static, Single \\
Our Method & PPO & CTDE & Cooperation & Static+Dynamic & Robot & Distance to Target & Dynamic, Single

\end{tabular}%
}
\caption{Summary of recent related work on multiple mobile objects using deep reinforcement learning. Note that the reward objective with the correct path in the robotic applications includes collision avoidance term.}
\label{tab:related-work}
\end{table*}


\deleted{Methods such as in \cite{tai2017virtual, pfeiffer2017perception}achieve single-source to single-destination planning with collision avoidance using neural network techniques. In the later research, research focuses on further improving existing models or solving different sub-problems. \citet{zhang2017deep} deployed deep reinforcement learning algorithms to handle new conditions in both simulated and practical environments; \citet{li2021dynamic} proposed a dynamic and scalable user-centric route planning algorithm based on polychromatic sets theory that can provide customised routes for users based on their specific requirements and various attributes of roads. \citet{zhang2021reinforcement} presented a reinforcement learning-based opportunistic routing protocol to address communication problems in an underwater acoustic sensor network, where the environment is with a high bit error rate, long delay, and low bandwidth. \citet{zeng2019navigation} considered an unknown and dynamic environment in which obstacles can dynamically move; \citet{zhelo2018curiosity} introduced an additional reward that leads to faster learning and better exploration ability; a modular method is proposed in \cite{wang2018learning}, where tasks are distributed into different DRL streams to accelerate the training, e.g., using two separate DRL agents, a spatial and a temporal Deep Q-Network (DQN), to perform local planning and global planning, respectively. For multi-source planning, the collision avoidance feature was enabled in a multiple robots environment \cite{chen2017decentralized,long2018towards}, where a DRL agent was trained using local information, e.g., Light Detection And Ranging (LiDAR) results, the goal information or the position of the goal. \citet{zong2022mapdp} proposed a DRL-based method to address Cooperative Pickup and Delivery Problem (PDP), which is an NP-hard problem and is hard to be solved optimally. \citet{tai2018socially} and \citet{jin2020mapless} demonstrated the multi-source planning approaches can be compatible with human pedestrians navigation in a highly dense crowded environment, where multiple sensors, e.g., LiDARs and depth cameras, are used to avoid end-to-end collisions \cite{liang2021crowd}. These methods mainly improve the training speed or avoid collision while moving, however, the target point is static and cannot be changed before reaching the destination. }

\added{Previous solutions tackle the DMS-SD problem using enumerated methods~\cite{xiao2022efficient}, which generate as many optimal or near-optimal solutions as possible without learning any patterns from past experience or only leveraging limited information from the earlier trajectories~\cite{xiao2023collaborative}. Both approaches require high computational resources (e.g., high CPU frequency) to rapidly compute the optimal results, while the resources are strictly limited on mobile devices.}

\added{A vast majority of existing navigation strategies use Multi-Agent Deep Reinforcement Learning (MADRL) and focus on guiding robots, intelligent vehicles, and Unmanned Aerial Vehicles (UAVs) with the aim of minimising the overall costs (e.g., energy consumption, time and distance). \citet{han2020cooperative} propose a DRL-based approach to cooperatively address the multi-robot navigation problem in a dynamic environment with multiple static targets and multiple mobile obstacles. \citet{long2018towards} propose a sensor-level decentralised multi-robot navigation framework, where the collected data from sensors are directly mapped to the collision-free objective in a shared policy, enabling the deployment of the same DRL agent across multiple scenarios. 
The proposed navigation system in \cite{liu2021visuomotor} utilises graph neural networks to extract key features from only the local motion information, enabling collaboration between multiple robots and, therefore, generating a mapless visuomotor policy.
\citet{mohseni2019interaction} introduce a decentralised learning strategy for multiple mobile robots with multiple individual stationary goals, where the local information between robots is not shared. 
The method in \cite{sivanathan2020decentralized} presents a decentralised multi-robot navigation framework, enabling robots to move towards their independent targets with an optimal path and collision-free. 
\citet{xue2023multi} construct a centralised training and decentralised execution (CTDE) multi-UAV navigation algorithm, integrating recurrent neural network into MADRL, which can coordinate UAVs to navigate safely in unknown and complex environments. 
The approach in \cite{yang2020autonomous} learns the local known environment, which allows multi-UAV to navigate and avoid mobile obstacles in a dynamic environment. 
\citet{moon2021deep} proposed a Cramér–Rao lower bound-based framework enabling multiple UAVs to track multiple First Responders (FRs) in challenging 3-D environments in the UAV-aided network. 
\citet{huang2022vision} designs a vision-based method that mainly utilises onboard vision sensors and inertial measurement to achieve collision avoidance for multi-UAV navigation. 
}

\added{Overall, as summarised in Table \ref{tab:related-work}, previous related studies mainly focus on collision avoidance and planning optimal path features for robots and goal-tracking features for UAVs. However, existing methods either only consider the static objectives or the objectives randomly move with their individual goals (e.g., tracking), which are independent of the moving object (e.g., robot or UAV). The goal generated by our proposed problem is dynamic, which changes regularly when a robot moves, and is highly related to the moving objects themselves. Moreover, the existing reward function and DRL algorithm combination cannot fulfil or perfectly fit the requirements of addressing the proposed DMS-SD problem, leading to a slow convergence rate and long training time. In this paper, our method complements the gaps by designing a new potential-based reward function and leveraging the masking mechanism to achieve a superior performance.}

%% file: assets/documents/04methodology.tex
\section{Problem Definition}
\added{In this paper, we consider a map as the aerial view of the factory layout, constructed by a rectangle with $X$ width and $Y$ length. The factory contains a set of static obstacles $N_s$, such as boxes and oil tanks, and dynamic obstacles $N_d$, such as other robots and humans. The factory deploys a set of mobile robots $N_r$ navigating on the ground plane to distribute and deliver materials from the warehouse to workshops. The objective of every information update is to select a rendezvous position to minimise the overall time and travelling distance to the future delivering tasks. However, the desired rendezvous distribution position can be frequently and dynamically changed when new information arises from robots, which requires a fast result computation speed. Additionally, robots should reach the rendezvous position without any collision with obstacles.}

\section{Methodology}

\begin{figure*}[ht]
    \centering
    \includegraphics[width=1\textwidth]{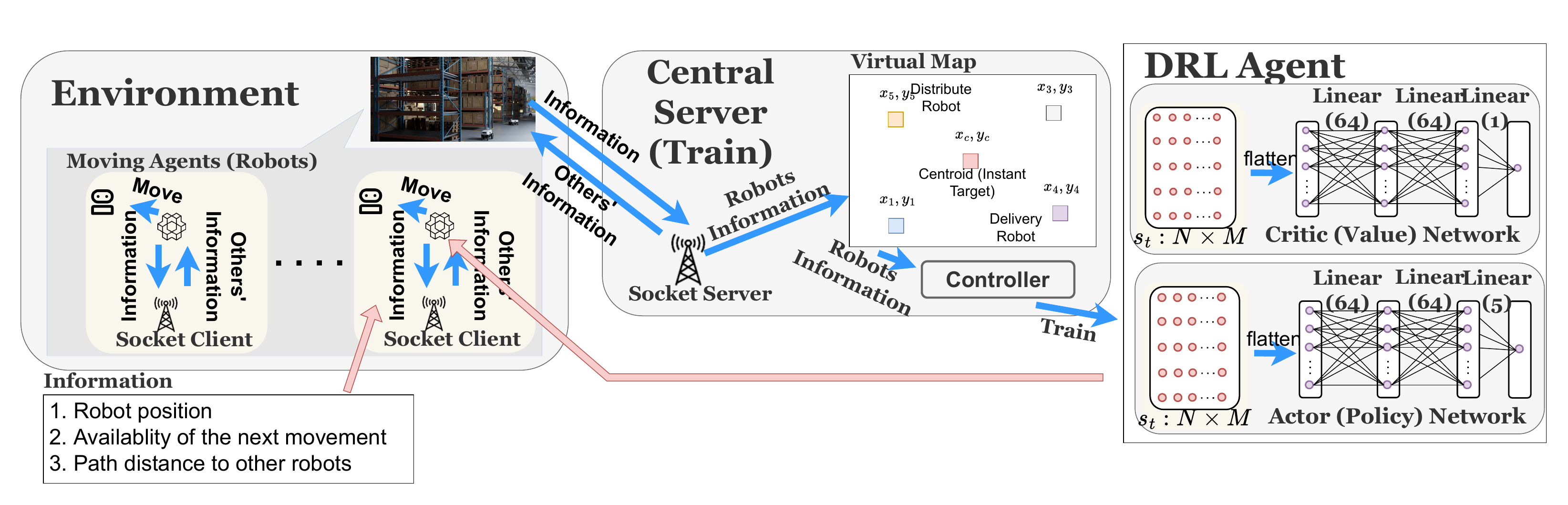}
\caption{The framework of our DRL-based DMS-SD solver}
\label{fig:workflow}
\end{figure*}

Fig.~\ref{fig:workflow} shows the framework of our DRL-based method. There are three main parts in the system: the \emph{environment}, the \emph{central server} and the \emph{DRL agent}. The \emph{environment} provides a place where robots can move. It accepts and applies future actions from the \emph{DRL agent}, so that the robots can make corresponding decisions based on the predicted actions. When the \emph{environment} has finished applying the accepted action, it responds to a current state $s_t$ and a reward $r_t$ to the \emph{DRL agent}. $r_t$ indicates if the training robot is approaching \deleted{to} an optimal solution \added{defined by the centroid of robots' positions}\deleted{, which is provided by the \emph{modified Dijkstra's algorithm}~\cite{xiao2022efficient} based on robots' current positions}. \deleted{The \emph{DRL agent} uses an actor neural network to make the predictions. and stores received information in the \emph{replay memory}. The \emph{optimiser} is used to manage the update of neural networks, which is much easier than manually updating weights.} The \emph{central server} is used to train the DRL agent and distribute the updated version to all robots. It also provides a platform to exchange critical information between robots. In the following subsections, we describe the critical components of this framework in detail.

\deleted{We make use of the widely accepted Proximal Policy Optimisation (PPO) algorithm~\cite{schulman2017proximal} for our DRL framework, where at the end of each \emph{timestep}, $t=1,2,\dots T$, an agent receives an observation of the environment's state, $s_t$, and provides an action, $a_{t+1}$, (the next movement) based on a learned policy $\pi_\theta$, and receives a reward $r_t$. PPO is an actor-critic style deep reinforcement learning algorithm, which can quickly converge to a stable policy via monotonic updates with a small amount of training samples. It contains two neural networks: the \emph{policy (actor) network} $\pi_\theta$ and the \emph{critic network} $V_\varphi$, where $V_\varphi$ evaluates the performance of the \emph{policy network}, and $\theta$ and $\varphi$ are the network parameters for the \emph{actor} and \emph{critic network}, respectively. Because PPO is an \emph{on-policy} algorithm, the agent collects trajectories (the entire path from a step $t_i$ until a complete signal at time step $t_j$) and stores it into the \emph{replay memory} for later use. During the training process, the agent samples a \emph{batch}, i.e., containing certain time steps from the \emph{replay memory} and splits it into multiple \emph{mini-batches}. Neural networks are updated by calculating the loss between predictions and real values in \emph{mini-batches}.}

\subsection{Environment}
\added{We build an environment to act as an intermediate component between the DRL agent and the factory.} The environment simulates the actual maps in the real world in which the agent can send actions and receive responses. Specifically, the environment contains three main steps to simulate the map: initialisation, time step action and done action. The initialisation reads a pre-defined map \deleted{and calculates the Adjacent Matrix(s) of all pairs in the map, which applies the modified Dijkstra’s algorithm~\cite{xiao2022efficient} multiple times.} The time step action is repeated until the goal is reached. Each time step action involves accepting actions from the agent and provides a response consisting of a tuple $\{s_t,r_t,done_t\}$, where $s_t$ is the current state described in Section \ref{st:state-space}, $r_t$ is the evaluation of the action described in Section \ref{st:reward}, and $d_t$ is marked as true if all robots are at the instant target position \added{(i.e., no further movement is needed to reach the goal)}\deleted{, the goal is achieved}. During each time step, robots' movements are controlled by the action. \deleted{The movements of each robot, namely a $trajectory$, including status during the move, are collected by the agent stored in the replay memory, which is forwarded to the $optimiser$ and trained using PPO.} More importantly, we simulate the $dynamic$ feature in which the goal is changed (recalculated) each time robots move one step - they should not move towards the old goal if they did not move as expected. When $done_t$ is true, where all robots are in the same position, the environment is reset and starts from the initialisation step again. 

\paragraph{Centralised Learning Process}
\added{We use a centralised strategy to train the DRL agent. As shown in Fig.~\ref{fig:workflow}, the information of each robot is broadcast to all other robots and temporarily stored in the \emph{Central Server}, which constructs a virtual map to simulate the factory layouts (i.e., training environment). The virtual map synchronises with the real environment or the simulator with critical information, such as actual factory layouts and obstacle information. The central server updates the DRL agent regularly and distributes it to each robot via socket connections. In the real deployment environment, the \emph{Central Server} is only used to broadcast their information, and each robot stores a well-trained DRL agent to predict its next movement.}

\paragraph{Reward Function}\label{st:reward}
At each step, the DRL agent sends an action to the environment, and the environment responds with an immediate signal to the DRL agent, which is the reward. The reward is a key part of an environment that can evaluate the input action, and the purpose of training a DRL agent is to maximise the reward. \deleted{Note that the design of a good reward function is significant to the success of a DRL method. Many existing studies, such as those in \cite{abdi2021novel,liu2020mapper}, simply use fixed values for each movement and achieve the goal. The main drawback of this selection can be that the agent is hard to find the 'first goal', which is the first time all robots meet at a place during the training. Specifically, we say achieving the first goal is nearly random using such a reward in \cite{abdi2021novel} because the goal is not navigated by reward values. For example, in \cite{abdi2021novel}, the proposed method penalises movements for a -1 score and receives 50 scores if it reaches the target. There is no reward difference for moving in any direction unless the goal is reached. This reward strategy works if there is only one robot, as the agent will eventually locate the target by exploring more unvisited nodes. However, for multiple robots, the probability of randomly finding the meeting destination drops significantly and a navigated reward is, therefore, needed.}

In our work, we use a potential-based reward function that calculates the difference in the potential from the current robots' position to the optimal. This means the reward is directly related to the objective in which maximising the receiving reward is the direction towards the optimal solution. More importantly, the reward value helps the agent to reach the goal, especially when there is a lack of information from the target due to \replaced{rapid}{rapidly} changing in the destination. 
We define the reward function $R$ as shown in Eq.~\ref{eq:reward}
\begin{equation}\label{eq:reward}
    R(s_t)=r + F(s_{t-1},s_{t})
\end{equation}
where \replaced{$r$ is a positive value rewarded}{$r=1$} $r$ is a positive value rewarded only if all guided robots have reached the goal and $r=0$ otherwise. 
$F$ provides the domain knowledge of transiting from the previous state $s_{t-1}$ to the current state $s_{t}$ which is defined in Eq.~\ref{eq:reward-f}
\begin{equation}\label{eq:reward-f}
    F(s_{t-1},s_{t}) = \begin{cases}
    r_1, &\phi(s_{t-1}) - \phi(s_{t}) < 0\\
    r_2, &\phi(s_{t-1}) - \phi(s_{t}) > 0\\
    r_3, &\phi(s_{t-1}) - \phi(s_{t}) = 0
    \end{cases}
\end{equation}
\added{where $\phi(s)$ is the potential function calculating the potential of a state $s$. The specific rewards $r_1, r_2$ and $r_3$ are defined as follows:}
\begin{itemize}
    \item \added{A positive step reward $r_1$ is applied for guiding the robot to reach or approach the goal, i.e., a positive difference of the two potentials.}
    \item \added{A extra-negative reward $r_2$ is used when the robot moves further away from the planned target, i.e., a negative difference of the potentials.}
    \item \added{A small negative step reward $r_3$ is applied when the robot remains static or moves to a position with the same distance as the previous position, i.e., the difference of the potentials does not change.}
\end{itemize}

\added{The potential $\phi(s_{t})$ is defined as the optimal path (excluding obstacles) distance to an instant target, which can be dynamically changed every step. The target is calculated by the centroid of all robots' positions. That is, the $x$ and $y$ coordinates in the target are the mean values of all robots. If the chosen target is an obstacle, we randomly select a neighbour instead to avoid an inaccessible route.}

\deleted{where $\phi(s)$ is the potential function calculating the potential of a state $s$, $r_1$ should be a positive number indicating a better change of the potential, $r_2$ penalises a worse transition, $r_3$ penalises the same potential value, and these rewards should satisfy $|r_1| < |r_3| < |r_2|$ to avoid detouring. In this paper, we define $r_1$, $r_2$ and $r_3$ as 1, -10, -5, respectively.
To identify the potential of a state, we first calculate the expected target}
\deleted{$target' = argmin(D)$}
\deleted{where $D$ is the distance between the current position to the target calculated by the modified Dijkstra's algorithm. The result $target'$ is one of the optimal destinations that all robots travel a similar distance to this point. Then, the potential function $\phi$ can be defined as the distance to $target'$, as shown in Eq.~\ref{eq:reward-phi}}
\deleted{$\phi(s) = -M_{adj}[s][target']$}
\deleted{where $M_{adj}[s][target']$ means extracting the shortest distance in $M_{adj}$ from a state $s$ (with some coordinates information) to another point $target'$. A negative value of the change in distance $\phi(s)$ indicates the distance becomes further away, and the robot is moving further, while a positive $\phi(s)$ means the robot is moving towards the target. Otherwise, the robot stays in the same position. Note that in our testing experiments, no penalty to staying causes a convergence problem, which slows down the learning process and only converges to a sub-optimal solution. Therefore, we penalise the staying with a smaller penalty than moving further.}

\paragraph{Action Space}\label{st:action-space}
The action controls the movement of each robot, which consists of four directions and a staying state $\mathcal{A} = \{up, down, left, right, stay\}$. Therefore, the action can be defined as $a = <\mathcal{A}_i>$, $\forall i \in \{1...I\}$, where $I$ is the size of $\mathcal{A}$ and $\mathcal{A}_i$ means applying the $i$th action in $\mathcal{A}$ to the robot. Since $i$ is a $discrete$ number, we use the policy network $\pi_\theta(i_t|s_t)$ to select an action index $i_t$ from a finite set of $i$, given observation $s_t$, which is encoded by the policy network using $softmax(\overrightarrow{z})_i = \tfrac{e^{z_i}}{\sum_{j=1}^K e^{z_j}}$ where $\overrightarrow{z}$ is the policy's output, $K$ is the number of elements in $\overrightarrow{z}$ and $softmax(\overrightarrow{z})_i$ calculates the softmax value of $i$th element $x_i$ in $\overrightarrow{z}$.

\added{Additionally, we leverage the masking strategy proposed in \cite{huang2020closer} to mask out invalid actions. We define the obstacle, either mobile obstacles (e.g., moving robots observed by cameras or sensors) or static obstacles (e.g., walls), on the next movement as the invalid action. Specifically, when an obstacle is detected, the corresponding $logit$ is set to an extra-small value, leading to nearly no chance of being selected as the next action. This significantly saves time in training a robot to avoid obstacles, particularly in real-world training where hitting an obstacle may lead to unexpected failures and an increase in training costs. }

Note that there is only one action is assigned to a robot using this strategy, but the movements of multiple robots should be allocated. Two options can be considered: the $multidiscrete$ action space and using the same policy network to control all robots. The first option considers an action list $a_l = <\mathcal{A}_{i_1}, \mathcal{A}_{i_2} ,...., \mathcal{A}_{i_N}>$, $\forall N \in \{1...n_p\}$, where each action in the action list assigns an action to the corresponding robot, e.g., $\mathcal{A}_{i_1}$ applies $i_1$th action in $\mathcal{A}$ to the first robot. However, in our testing, this strategy takes a longer time to learn the correlation between robots, which is unnecessary. More specifically, the moving pattern for each individual robot should be identical\deleted{,} because they are all moving towards the same goal. In the $multidiscrete$ perspective, each robot holds \replaced{its}{their} own model (partial output of the neural network), which leads to a longer training time, although the individual model will be very similar eventually.
The second option applies the same policy network to each robot. However, from the environmental point of view, there is only one training robot. That is, the input and the output are calculated based on the first robot among all robots, as we described in the previous paragraph. Other robots do not return any information to the environment; instead, they only invoke $\pi_\theta$ with their corresponding state, which provides only an action index $i$ to the invoking robot. This procedure does not involve any training and, therefore, has no effect on the training agent or the training robot. As a result, only one neural network model will be trained, and this is the main reason why our approach shows competitive training speed.

\paragraph{State Space}\label{st:state-space}
The state space is the observation of the current time step. It provides useful information to the neural network so that the network can extract and learn patterns from the provided data. Our state space $S$ involves two parts: the positions of each robot in a coordinated form, and the sum distance from the currently selected robot to all other robots. In the first part, we considered adding the entire map as a part of the state, and it provides redundant information to the neural networks, slowing down the agent training. Since PPO is a model-free method, it can learn map information during the training; therefore, the up-to-date locations of each robot provide\deleted{s} enough \deleted{and} unique information to PPO's neural networks. More importantly, the feature extraction layer can be quickly trained because of the accurate information. The second part of the state eases the calculation on the neural network side because it is directly related to locating the goal point. Additionally, the distance is calculated by the actual distance (the shortest step taken) rather than the Euclidean distance. This is to avoid the local optimal solution when there are barriers. Overall, the neural network should be able to estimate the position of the goal based on the provided information and make a correct (highest reward) decision.

\added{Note that our approach aims to adapt different numbers of robots. That is, the model is trained within $n_p$ robots and can accept 2 to $n_p$ robots in the deployment. Specifically, at each resetting cycle during the training, the number of robots is reset to a random number with the range [2, $n_p$]. For those training with less than $n_p$ robots, zero padding is used to maintain the state space consistently. }

\paragraph{Neural Networks}
Before the observation (state) is sent to the policy and critic network, we use a Multi-Layer Perceptron network to extract features. This network contains two layers, each layer has 64 neurons and uses $Tanh$ as the activation function, which can output a latent representation for the policy and a value network. \added{We use the same neural network settings as shown in a well-known DRL library \cite{stable-baselines3}.}


\subsection{Decoding Real-World Map And Communication}
We translate the real-world production map into a virtual grid environment based on vertices and edges, enabling the DRL agent to guide each robot towards a shared target via this environment. The DRL agent only requires the position of each robot; thus, the initial step involves transforming the real-world map into a grid map to delineate accessible paths and obstacles. Specifically, we generate a grid of dimensions $X \times Y$, covering the entire map, where $X$ and $Y$ denote the length and width of the map, respectively. This grid represents the vertices in the previous descriptions, allowing robots to move solely between adjacent grids (vertices). \deleted{However,} \replaced{C}{c}ertain vertices become inaccessible due to static elements (e.g., walls, doors, and installed machines) and dynamic obstacles (e.g., moving people and other robots). Those inaccessible movements are masked out by the masking strategy to achieve the obstacle avoidance. \deleted{This paper focuses exclusively on static obstacles, leaving the precise identification of dynamic obstacles to local planning algorithms. We designate these static obstacles as impassable, removing edges connected to adjacent vertices to prevent the DRL agent from planning routes through these vertices. Subsequently, the converted grid-based map is introduced into the training environment, and the corresponding adjacency matrix is computed.}

Meanwhile, synchronising robots' information is imperative to prevent unreliable predictions. Consequently, the subsequent phase involves facilitating communication among robots to update their current states. Instead of using the centralised mode, we adopt a distributed approach: each robot independently stores the DRL model and regularly updates it. The adoption of a distributed model is advantageous for our problem due to (i) enhanced fault tolerance, where a robot's (e.g., robot) failure does not impact other robots; (ii) scalability, enabling the deployment of more robots as long as they join the local network; (iii) reduced latency, as local processing circumvents the need for server requests, leading to delayed result generation; (iv) greater flexibility, as the hardware specifications of a robot's machine (e.g., Raspberry PI 3B) can be significantly lower than that of a centralised server (e.g., i9-10900K), thereby conserving energy and promoting environmental sustainability. Consequently, rather than transmitting information to a central server, each robot broadcasts its state to all other robots\deleted{, including the current location (i.e., grid position)} acquired from local sensors. The latest local version of the robot states is then relayed to the local copy of the DRL agent. Following the policy network process, a discrete number is generated, subsequently interpreted by the environment and relayed to the robot, guiding it to the next vertex.

%% file: assets/documents/05evaluation.tex

\section{Evaluation}
In this section, we evaluate the performance of our DRL method to solve the DMS-SD problem in material distribution. We prove that our DRL method can reach an optimal or near-optimal solution when the model is well-trained\deleted{,} and can also converge fast and respond to the new navigation request in a very short time, at the millisecond level. This is critical because rapid access is significant when robots are directed by the intelligent control system in real-time navigation, which requires low latency (i.e., fast responding time). 
 
\added{We have conducted experiments by training models under a varying number of robots from 2 to 12 (in the same model using zero padding strategy) in different sizes of maps from 50$\times$50 to 70$\times$70, respectively. The map is used to virtually simulate the different factory layouts. The factory is split into squares (each equivalent to 1 square meter in our simulation), as shown in Fig.~\ref{fig:example}. Robots may move within the map with a uniform speed and can only move 1 square each time.} These maps are randomly generated and simulate the environment using \emph{NVIDIA Isaac Sim}\footnote{https://developer.nvidia.com/isaac-sim}, which includes obstacles, paths and robots. Specifically, we generate an experimental map in four steps. Firstly, an empty map is generated, where a node is connected to all its adjacent nodes. Then, obstacles are randomly \replaced{placed}{added} on the map, \replaced{which includes both static and dynamic obstacles. For each map node, there is a $5\%$ probability of generating a static obstacle (i.e., static obstacle density), whereas there is an additional $2\%$ chance of generating a mobile obstacle (i.e., dynamic obstacle density), which can randomly move in the map. }{but we are not building a maze, where obstacles are only used to ensure our method can deal with static barriers}. Thirdly, we build workshops to accept the material delivery and randomly assign initial positions to robots. Finally, before the experiment starts, the generated map is examined to ensure robots can eventually meet at a point when robots are at any initial positions. 

Each model is trained with 150 iterations. Other DRL agent parameters are held constant as: the GAE learning rate $\gamma=0.99$, the variance control parameter $\lambda=0.92$, the clip range $clip=0.2$, the optimiser's learning rate of 0.001 (Adam Optimiser~\added{\cite{kingma2014adam}}), and the \emph{entropy coefficient} of 0.001. \added{The default parameters of our approach in the implementation are set as follows: $r_1 = 1,r_2 = -10$ and $r_3 = -5$.}

\deleted{To prove the navigation ability of the proposed method, the real-world experiment simulates the factory scenario by separating the factory layout as a 15$\times$15 size map. In experiment, three robots navigate from the initial locations to the encounter point at the same time along routes calculated by the algorithm. Simultaneously, the experiment evaluated the well-trained model could direct robots to navigate effectively wherever initial locations of robots. The experiment result support the further practice in the real factory environment. }


\begin{figure*}
\begin{multicols}{2}
     \centering
     \begin{subfigure}[b]{0.49\textwidth}
         \centering
         \includegraphics[width=\textwidth]{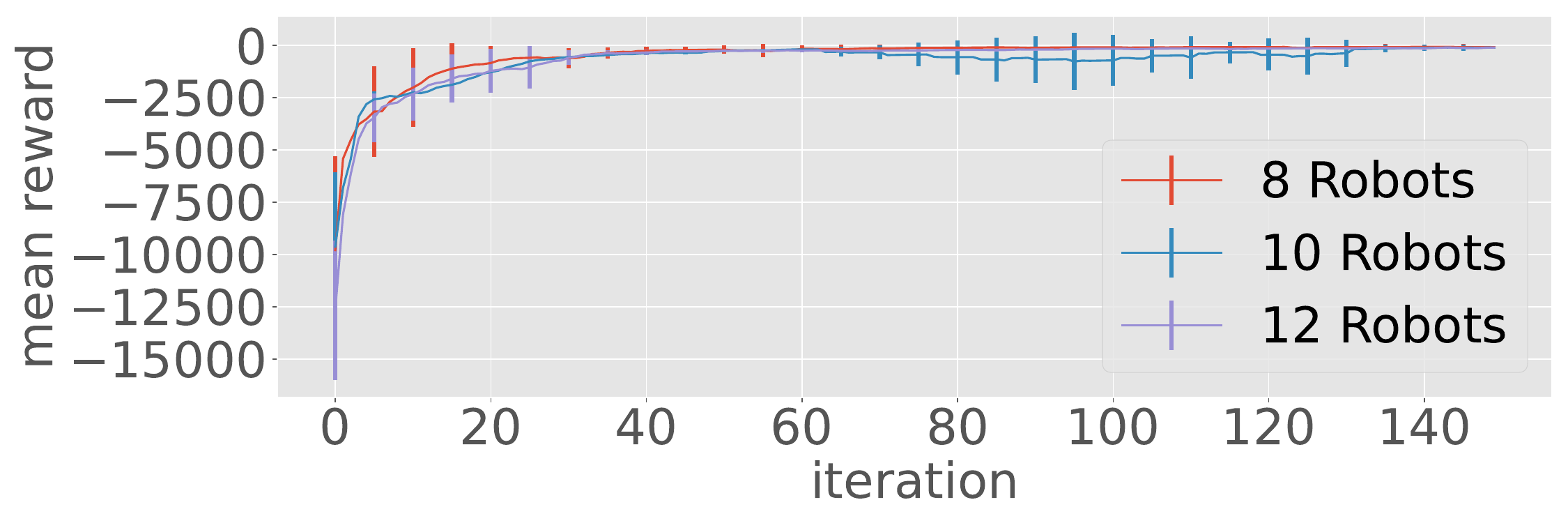}
         \caption{RM for the different numbers of robots.}
         \label{fig:rm-rbs}
     \end{subfigure}
     \begin{subfigure}[b]{0.49\textwidth}
         \centering
         \includegraphics[width=\textwidth]{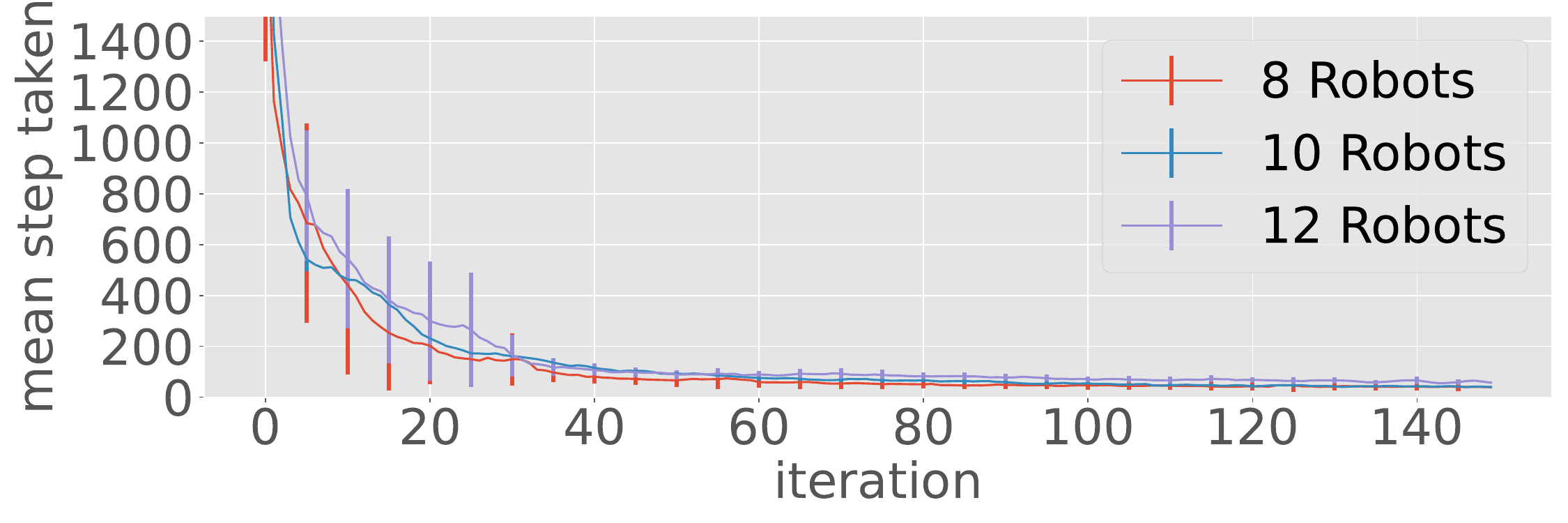}
         \caption{MST for the different numbers of robots.}
         \label{fig:MST-rbs}
     \end{subfigure}
     \begin{subfigure}[b]{0.49\textwidth}
         \centering
         \includegraphics[width=\textwidth]{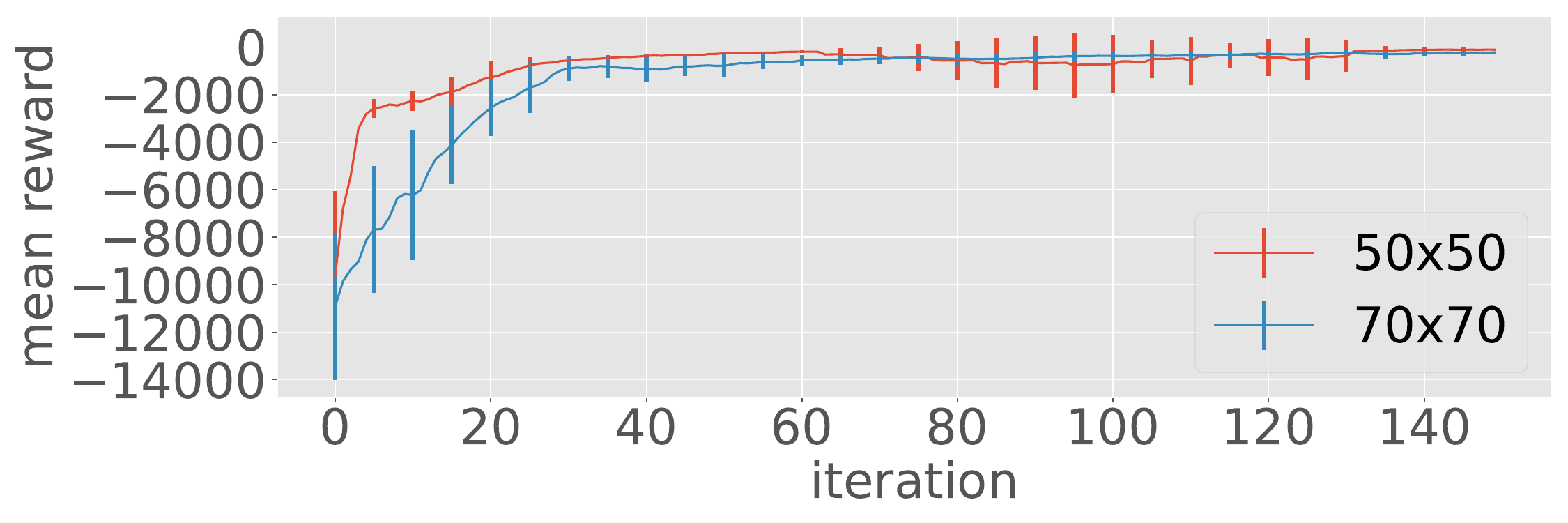}
         \caption{RM for the different sizes of the map.}
         \label{fig:rm-sm}
     \end{subfigure}
     \begin{subfigure}[b]{0.49\textwidth}
         \centering
         \includegraphics[width=\textwidth]{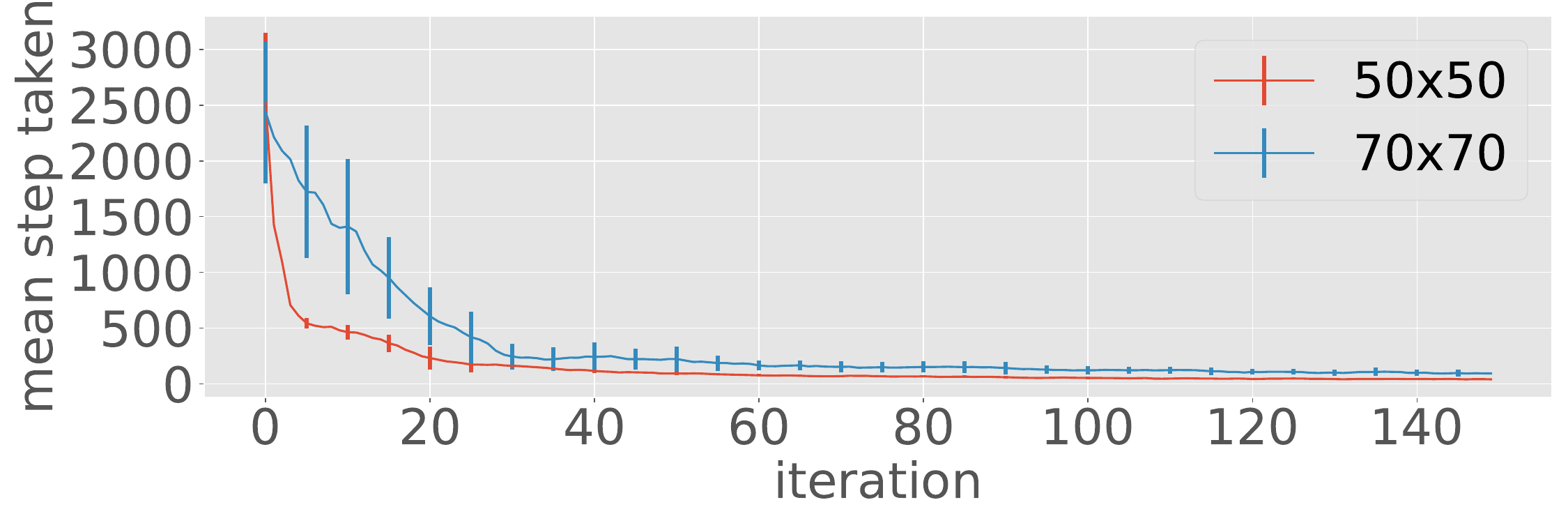}
         \caption{MST for the different sizes of the map.}
         \label{fig:MST-sm}
     \end{subfigure}
\end{multicols}
\caption{Training performance of different numbers of robots and map sizes.}
\label{fig:rm-MST}
\end{figure*}

\paragraph{Remark 0: Mean Step Taken (MST) and Rewards Mean (RM)} 
Two evaluation metrics are used to analyse the performance of our DRL method: (i) Steps Taken Mean (MST), and (ii) Rewards Mean (RM) \cite{stable-baselines3}. Instead of simply focusing on the statistics of each time step, we particularly use episodic information to evaluate the overall performance of our DRL method. MST is the average time steps consumed for an achieved goal within an iteration. The number of time steps can be accumulated to the next iteration due to no $done=True$ response, which means the goal has yet to be achieved, i.e., robots still need \added{more steps} to be met at the same place. In addition, RM is counted similarly to the MST, but it collects information on the agent's reward. \added{Fig.~\ref{fig:rm-MST} shows the MST and RM performance in different numbers of robots and map sizes. All experiments are performed 3 times, and the corresponding standard deviation is calculated as the error bar, indicated as the vertical line.}

\added{Results of MST and RM on a 50$\times$50 size map with different numbers of robots are shown in Fig.~\ref{fig:rm-rbs} and Fig.~\ref{fig:MST-rbs}. It is noticed that all trained agents achieve a stable MST number after around 100 iterations of training. Although they all converge fast to the optimal agent, fewer robots lead to slightly faster convergence speed due to low path planning complexity to the optimal solution.
Although all trained models eventually reach an extremely low MST number ($\leq 40$), the best MST with a larger number of robots is slightly higher. For example, the MST in the 12 robots case is around 37, which is higher than the 8 robots case (around 34). This is because there may be a few more steps needed to wait at the same place for other robots to achieve the target.
Spikes shown between iteration 80 and 110 indicate that the DRL agent is trapped in a local optimal in which the DRL agent is temporarily difficult to learn and adapt to new conditions. However, the agent rapidly escapes from the local optimal and backs to the global optimal within a few iterations. }

\added{Similarly, results of MST and RM in various size maps with 5 robots are shown in Fig.~\ref{fig:rm-sm} and Fig.~\ref{fig:MST-sm}. MST results demonstrate that the proposed DRL method converges steadily to be less than 40. The 70$\times$70 map size case requires more than 120 iterations to converge to a low MST, while the 50$\times$50 case only needs less than 100 iterations. Since the positions of robots are randomised when the environment is reset, there are more available routes to be explored on the larger map, which may result in being trapped in a local optimal, such as peaks at around the 95 iteration. Thus, the agent needs a longer time to learn a better route. Note that although MST can approach a considerably low value ($\leq 80$) in the 70$\times$70 map size case, it requires more iterations to reach a near-optimal solution. This is because the training agent tries to learn details of the optimal path while simply choosing the correct direction of the path takes fewer training iterations.}

Another metric, RM, is expected to show a similar trend but \added{a} reversed tendency. It is noticed that the converged reward is less than the expected value. Specifically, according to the reward function Eq.~\ref{eq:reward}, an optimal value should be a positive number that is equivalent to the number of movements. This means there are unexpected movements, which could be caused by the same reason as above: the training agent is waiting for other robots to reach the destination, where the waiting is penalised. Therefore, the reward can be less than expected if the training agent takes the correct steps to reach the goal with a few steps of waiting. Moreover, RM is calculated based on \added{each iteration}, and the chance of waiting for other robots is more frequent, thus lower values of the reward will be dominant, leading to a low average value.

\added{Overall, these results demonstrate that our method is compatible with various numbers of robots or map sizes. More importantly, the trained agent is able to fit multiple numbers of robots in the same model. For example, with a trained model of 12 robots, it is able to fit 2 to 12 robots by adding zero padding for those state spaces less than the predefined size, i.e., 10 robots implementation should append 4 zeros after every state $s_t$.}


\deleted{Results of MST and RM on a 20$\times$20 size map with different numbers of robots are shown in Fig.~\ref{fig:results_b}. It is noticed that all trained agents achieve a stable MST number after around 800 episodes of training. Fewer robots lead to fast convergence speed due to low path planning complexity to the optimal solution, for example, the MST converges fast under the 2 robots case, where MST drops significantly in the first few episodes. 
Notably, in more robots' cases, the MST becomes more stable, where the MST curve only fluctuates within a minor range. This can be shown in the 2 robots case that there are spikes at 1,300, 2,000, 2,400 and 3,700 episodes, while 6 and 10 robots are more stable when the model has been well-trained, e.g., after 2,000 episodes. The reason could be that robots share the same neural network model, which leads to less chance of having multiple solutions for a single time step. Thus the model does not need to learn the reward of other optimal options. Furthermore, spikes in the 2 robots case indicate that the DRL agent is trapped in a local optimal in which the DRL agent is temporarily difficult to learn and adapt to new conditions. However, the agent rapidly escapes from the local optimal and backs to the global optimal within a few episodes. This demonstrates that our method is more stable when there are more robots, but it is also compatible with a small number of robots.
Although all trained models eventually reach an extremely low MST number ($\leq 20$), the best MST with a more number of robots is slightly higher. The MST in the 10 robots case is around 17, which is higher than the 6 robots case (around 14), and the 2 robots case of 7 MST. This is because there may be a few more steps needed to wait at the same place for other robots to achieve the target. }


\deleted{Similarly, results of MST and RM in various size maps with 5 robots are shown in Fig.~\ref{fig:results_c}. MST results demonstrate that the proposed DRL method converges steadily to be less than 20. The 80$\times$80 map size case requires more than 3,400 episodes to be converged to a low MST, while other cases only need less than 1,000 episodes. Since positions of robots are randomised when the environment is reset, there are more available routes to be explored on the larger map, which may result trapped in a local optimal, such as peaks at around 2,000, 2,300 and 2,600 episodes. Thus, the agent needs a longer time to learn a better route. Note that although MST can approach a considerably low value ($\leq 20$), it requires more episodes to reach a near-optimal solution. This is because the training agent tries to learn details of the optimal path while simply choosing the correct direction of the path takes fewer training episodes.}

\paragraph{Remark 1: Computation Time} \deleted{Since our DRL agent is trained offline, t}The computation time \added{of our approach} only reflects the time cost of the states passing through neural networks, which can be very minor. It usually takes around \emph{30ms} in a desktop environment and less than \emph{100ms} on a mobile device (e.g., Intel i5 9600K macOS 13.2 v.s. Raspberry Pi 4 Model B, Ubuntu 20.04) because the defined neural network is not complicated. \added{More importantly, the input of our approach is nearly independent of the map size (only related to the maximum coordinates). That is, although the training time on a large map increases due to the exploration of the map, the actual computation time on a deployed device is minimally affected. This is the main reason that our method indicates a faster computation speed than enumerated methods. In contrast, methods in \cite{xiao2022efficient, xiao2023collaborative} are largely impacted by the map size, which can even increase to many seconds. This is intolerable for robots during real-time dynamic navigation and can significantly affect the overall efficiency of the factory.} \deleted{More importantly, the trained model is only related to the state of robots instead of map information. We observe the computation time of the modified Dijkstra's algorithm can reach up to 20s in a $80\times 80$ map, where the computation time is proportional to the map size and the number of robots.} 

\begin{figure}
    \centering
    \includegraphics[scale=0.52]{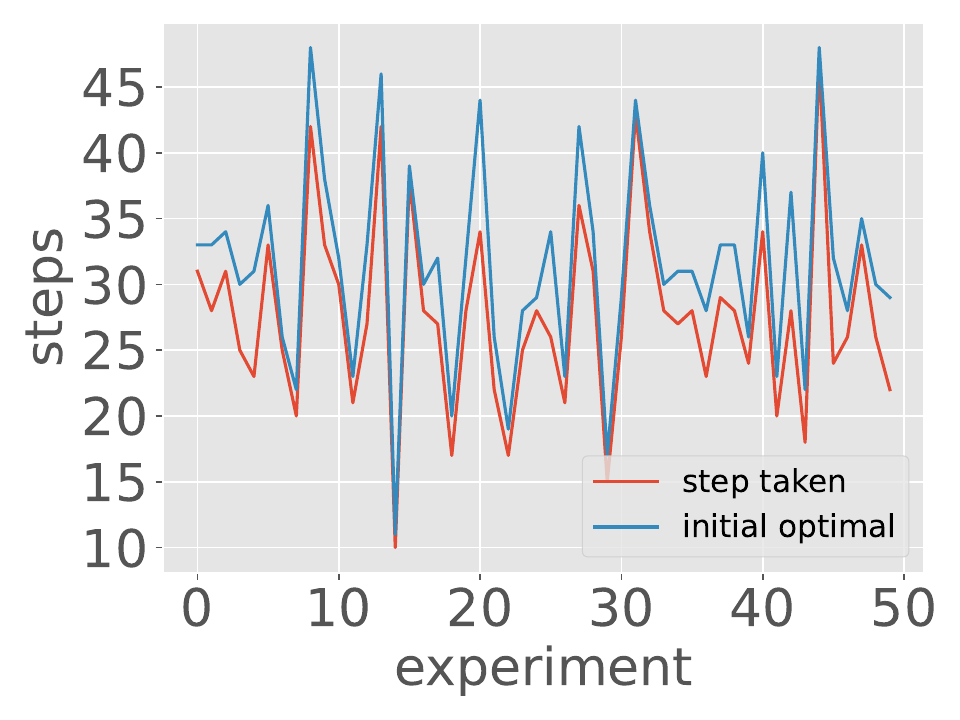}
\caption{Performance of our model v.s. optimal during the training on a 50x50 size map with 10 robots.}
\label{fig:optimal-results}
\end{figure}

\paragraph{Remark 2: Optimality}
We particularly observe the gap between our well-trained agent and the optimal solution. Fig.~\ref{fig:optimal-results} shows \added{the model test results of the time steps taken by a trained agent from the Remark 0 and the optimal step to the initial goal on a 50x50 size map with 10 robots. The \emph{x-axis} indicates the index of the experiment performed, and the \emph{y-axis} shows the number of steps cost.} \deleted{Both agents have trained 1,000 episodes, and the \emph{x-axis}, the iteration number, indicates the number of complete \emph{trajectories}, i.e., when $done=True$.} \added{It is obvious that the gap between the actual and optimal steps taken is minor, which is mostly less than 5 steps.} \deleted{The distance to the optimal target after the same range is around two, which is an acceptable distance. }
Notably, the optimal target uses the initial target that is calculated based on the initial positions of robots. However, we note that the final destination could be significantly different from the initial calculation, even if every movement is optimal. \added{This is because (i) there could be multiple solutions for a single state due to the objective: equal travel distance. Specifically, at every state, each robot may have multiple choices to reach the optimal goal, and the combination of different optimal choices may result in deviating from the initial target. Therefore, the final target may differ from the initial target, but both can still be optimal; (ii) the training agent has reached the goal, but it must wait for other robots to be at the goal, which could take a few steps; (iii) the instant target is based on the centroid, that is, there may be obstacles on the path to the target, leading to a detouring. This also changes the initial target.}
Overall, the performance of our method in this test is close to the optimal result, although there is a minor gap.

\paragraph{Remark 3: DRL Training Time}
It is evident that our method effectively decreases the MST to a considerably low level in the first several iterations, with a minor influence from the number of robots. According to the convergence of MST, we can arguably calculate the value of DRL training time, and usually, every iteration consumes less than 20 seconds for training. For example, in Fig.~\ref{fig:MST-rbs}, the MST converges fast with a different number of robots varying from 8 to 12 on a \added{50$\times$50} size map case. Specifically, all agents are able to converge steadily to the optimal solution within approximately 150 iterations, and therefore, the training time is within 3,000 seconds (maximum). Fewer robots lead to fast convergence speed and slightly less DRL training time with low path planning complexity to the optimal solution. \added{For example, the 12 robots case requires more than 140 iterations (2,800 seconds equivalent) to train, whereas MST in the 8 robots case has converged within 60 iterations, although there are spikes in the later training. The main reason can be locating a common destination and moving to it for more robots is challenging to the DRL agent, especially during the first couple of training iterations where the neural networks have none or limited knowledge about the map.}

\added{Similarly, in the left graph in Fig.~\ref{fig:MST-sm}, both MST performance of the agent trained on a 50$\times$50 size map converges within 80 iterations on various map sizes under 10 robots, showing the training time to reach a low level of MST is less than 1,600 seconds, which is extremely fast. However, the agent trained on a 70$\times$70 size map requires a longer time, approximately 110 iterations (2,200 seconds equivalent), to approach a near-optimal solution. Although the map size does not directly and significantly affect the state and the reward function, the DRL agent still needs more time to explore static obstacles on the larger map and requires more movements to the goal position on average.
Overall, either increasing the map size or the number of robots can increase the training iteration and the actual time spent on the training. }




\deleted{Remark 4: DRL Convergence Problem}

\deleted{We note that the agent may learn 'bad' states and output 'bad' actions during the training, resulting in a low reward and moving toward the wrong direction. This can be because the agent is trapped in the local optimal that the agent performs well locally but receives a low reward. This is a general issue of the DRL algorithm, and usually, the agent can escape from the local optimal after less than 100 episodes. However, in some extreme cases, the agent keeps learning bad states and never escapes from the local optimal. Entropy is a measure of randomness in a probability distribution and is used in reinforcement learning to ensure that the agent cannot be stuck in a sub-optimal policy. In PPO, the entropy loss is defined as the negative entropy of the policy, which is a measure of the amount of randomness in the actions taken by the agent. The entropy loss term is added to the optimisation objective to encourage the policy to be more stochastic and explore different actions. By increasing the entropy of the policy, the agent is more likely to try new actions, which can lead to discovering better policies. Therefore, we could slightly increase the coefficient of the entropy loss ($\leq 0.01$) in PPO to help the agent explore more unvisited nodes and prevent being trapped in the local optimal. }

\begin{table}[]
\centering
\begin{tabular}{lll}
Source & Reward Function & DRL Algorithm \\ \hline
\cite{han2020cooperative} & $\begin{aligned} \begin{cases} 5, & \text{arrives the goal}\\ 1+d_g^{t}, & \text{close to the goal}\\ 10*(d_g^{t-1} - d_g^{t}), & \text{otherwise}  \end{cases}\end{aligned}$ & PPO \\ \hline

\cite{liu2021visuomotor} & $\begin{aligned} \begin{cases} -\frac{10}{N_m}, & \text{move towards the goal} \\ -\frac{10}{N_m}-1, & \text{collision}\\ 20-\frac{10}{N_m}, & \text{aligns to the optimal path}\\ 10-\frac{10}{N_m}, & \text{arrives the goal} \end{cases}\end{aligned}$ & Double DQN \\ \hline

\cite{yang2020autonomous} & $\begin{aligned} \begin{cases} 25, & \text{arrives the goal}\\ -300, & \text{collision}\\ -1, & \text{movement}  \end{cases}\end{aligned}$ & PPO 
\end{tabular}%
\caption{Summary of reward function settings from previous work. $N_m$ is the maximum moving step, and $d_g$ is the distance to the goal. For \cite{han2020cooperative}, a penalty=-10 is applied if the robot collides with obstacles or penalty=0 otherwise.}
\label{tab:exp-reward}
\end{table}

\begin{figure}
\centering
 \includegraphics[width=0.44\textwidth]{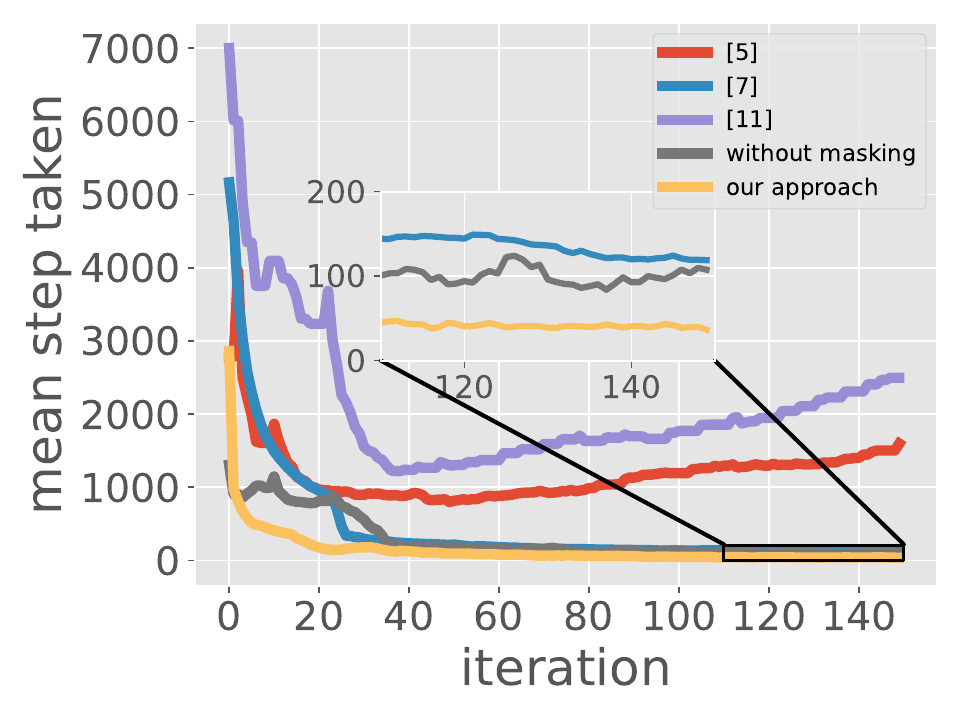}
\caption{Performance comparison of existing reward functions. `without masking' uses our approach without masking mechanism.}
\label{fig:rew-compare}
\end{figure}

\paragraph{Remark 4: Reward Functions}
\added{We list three reward functions proposed in the related work in Table \ref{tab:exp-reward} and use the DRL algorithm with a corresponding reward function to train the DRL agent. We use the same map with a 50x50 size map and 10 robots for the training and a random seed of 200 to ensure agents and the environment are fairly randomised. The MST results of those approaches are shown in Fig.~\ref{fig:rew-compare}. It is obvious that our approach performs the best among the 5 reward settings. Settings in \cite{han2020cooperative} and \cite{yang2020autonomous} cannot converge or even approach the optimal solution, whereas the setting in \cite{liu2021visuomotor} requires further training iterations to achieve a better result. The main reasons can be (i) the DRL agent has to spend a large number of iterations to learn obstacle avoidance by assigning a large penalty, whereas the masking mechanism marks the direction of the obstacle as an invalid action, which does not require any training.  (ii) the reward function may be too complicated to learn the patterns. Our designed reward function shows better convergence speed and is closer to the optimal solution, even without the masking mechanism. (iii) those previous reward functions aim to address the problem with static objectives while our goal constantly changes.}

\deleted{Remark 5: Reward Functions}

\deleted{We use a multi-agent strategy in this remark, where the input to the environment is a multi-discrete number, guiding each robot to the corresponding direction. Recall the action defined in our method is $a = <\mathcal{A}_i>$ (Section.~\ref{st:action-space}), we define a multi-discrete action as $a = <\mathcal{A}_{i_0}, \mathcal{A}_{i_1}, ... ,\mathcal{A}_{i_{n_p}}>$. We test three reward functions in this paragraph on a 20x20 map.} 

\deleted{Firstly, we extract two different reward functions from \cite{liu2020mapper} and \cite{qie2019joint} to compare the performance with our proposed reward function. The reward function developed in \cite{liu2020mapper} uses fixed rewards, which receive a -5 reward on collisions and a -0.1 reward on movements, whereas the latter one deploys a dynamic reward, calculating the distance from the robot to the target. \cite{liu2020mapper} is hard to locate the target when the number of robots becomes large because the reward function does not directly navigate robots to reach the meeting point. Specifically, the probability of robots' meeting at the same point is $\frac{1}{(n_{v})^{n_{p}}}$, which increases exponentially when either the number of robots $n_{p}$ or the number of vertices $n_v$ increases. This means robots can only meet at a common destination if they are all 'lucky' enough. As a result, the DRL agent is not able to learn and reward at the goal point - it assumes there is no goal (never ends) before the meeting. However, the training agent can learn the optimal solution when there are only two robots on a small map.}

%% file: assets/documents/06conclusion.tex
\section{Conclusion}
In this paper, we propose a highly effective and practical deep reinforcement learning-based method to address the DMS-SD problem. Our method shows high convergence speed during the training and can rapidly converge to the near-optimal solution. The method involves a novel reward function based on the outcome of the \replaced{centroid}{modified Dijkstra's algorithm}, which provides a simple strategy to help the agent learn the map information and speed up the training process. In the evaluation section, we demonstrate that reward functions from existing multi-agent systems may cause the convergence problem in our problem, which is far away from the optimal solution, \added{and our proposed approach shows superior performance than those approaches}. 
